\documentclass[journal]{IEEEtran}

\usepackage{cite}
\usepackage{amsmath,amssymb,amsfonts}
\usepackage{algorithm}
\usepackage{algpseudocode}
\usepackage{graphicx}
\usepackage{textcomp}
\usepackage{booktabs}
\usepackage{bm}
\usepackage{url}
\usepackage[hidelinks]{hyperref}

\graphicspath{{figures/}}

\begin{document}

\title{Sigma-Branch: Hierarchical Single-Path Network Reconstruction for Dynamic Inference with Reduced Active Parameters}

\author{Kohga~Tanaka~and~Hiroaki~Nishi%
\thanks{This work was supported by the JSPS KAKENHI (Grant Number JP26K02884). The authors also gratefully acknowledge support from the JST SIP Project (Grant Number JPJ012207).}%
\thanks{K. Tanaka is with the Graduate School of Science and Technology, Keio University, Kohoku-ku, Yokohama 223-8522, Japan (e-mail: tanaka@west.sd.keio.ac.jp).}%
\thanks{H. Nishi is with the Department of System Design, Faculty of Science and Technology, Keio University, Kohoku-ku, Yokohama 223-8522, Japan.}}

\markboth{Tanaka \MakeLowercase{\textit{et al.}}: Sigma-Branch: Hierarchical Single-Path Network Reconstruction}%
{Tanaka \MakeLowercase{\textit{et al.}}: Sigma-Branch: Hierarchical Single-Path Network Reconstruction}

\maketitle

\begin{abstract}
  Deploying deep neural networks on memory-constrained edge accelerators is bottlenecked by per-inference off-chip weight
  transfer rather than computation: the dense network cannot be
  retained on-chip, and every parameter must be loaded for every
  input. Existing model compression reduces this transfer only at the
  cost of permanent capacity loss. We propose Sigma-Branch (\(\Sigma\)B),
  a framework that restructures a pretrained dense network into a
 hierarchical binary tree composed of a shared backbone, hierarchical
 routers and specialized leaves. Pretrained weights are distributed
  across the tree via activation-based spherical \(k\)-means clustering,
  which jointly initializes router weights and per-branch channel
  allocations; soft-routing fine-tuning then aligns each leaf with its
  routed input subset. At inference, the resulting network executes only
  a single root-to-leaf path, reducing the active-parameter footprint
  while storing the complete dense parameter set in memory. Across CIFAR-100 / ResNet-50, ImageNet-1K / ResNet-50, and ModelNet40 / PointNet++, \(\Sigma\)B-Net reduces per-inference active parameters by 58--60\,\%
  while remaining within 1.72\,percentage points (pp) of the dense baseline Top-1. At
  comparable ImageNet-1K Top-1, the active-parameter reduction exceeds
  static structured pruning (FPGM, HRank) by 14--23\,pp. The
  cross-modal evaluation, spanning 2D vision and 3D point-cloud
  backbones, substantiates a framework-level claim that decouples
  per-inference memory traffic from the total parameter count.
\end{abstract}


\section{Introduction}\label{sec:introduction}

\subsection{Background}\label{sec:intro:background}

\subsubsection{Scaling of Deep Learning Models}

Deep neural networks (DNNs) have advanced markedly through scaling:
larger models trained with more data and compute consistently achieve
higher accuracy, from convolutional backbones such as
ResNet~\cite{he2016resnet} to Vision Transformers (ViT) of several
hundred million parameters~\cite{dosovitskiy2021vit}. These models
are designed under the assumption of data-center-grade compute and
are not natively suited to environments outside that regime.

\subsubsection{Rising Importance of Edge AI}

Concurrent with this scaling trend, an increasing share of inference
must be performed at the edge of the network, on devices ranging from
mobile phones to FPGAs and microcontrollers~\cite{chen2019deep}, for
reasons that include low latency, on-device privacy, and network-free
operation. The compute and memory capacity of such platforms are orders of
magnitude smaller than those of data-center systems, and inference typically
runs at a batch size of one, which together make direct deployment of most large-scale models impractical. Vision models for the edge, therefore, remain predominantly CNN-based (e.g.
ResNet~\cite{he2016resnet}, MobileNet~\cite{howard2017mobilenet},
ConvNeXt~\cite{liu2022convnext}) or MLP-based for 3D
data~\cite{qi2017pointnet,qi2017pointnetpp}, and even these backbones
continue to scale up in pursuit of accuracy, intensifying the
compute and memory demands placed on edge devices.

\subsection{Problem Statement}\label{sec:intro:problem}

\subsubsection{Memory-Constrained Accelerators on the Edge}

Modern DNNs exhibit a \emph{static} computation structure: every
parameter is typically accessed for every input, regardless of input
difficulty. As a result, each inference requires loading the entire
set of weights from memory for models that exceed the on-chip memory capacity.

Meanwhile, the edge increasingly hosts accelerators beyond GPUs---FPGAs,
NPUs, Edge TPUs, and microcontrollers---where on-chip memory (e.g., BRAM, SRAM) is orders of magnitude smaller than that of data-center systems.
Most practical models do not fit entirely on-chip, making weight loading
from off-chip DRAM a critical bottleneck on these platforms.

On such memory-constrained devices, weight loading directly governs inference latency: the volume of weights loaded per inference is limited by available memory bandwidth and on-chip capacity. Recent work has shown
that for large DNNs at small batch sizes, weight loading time dominates
while compute units remain largely idle~\cite{pope2023inference}. Edge GPUs
face the same issue: for example, the NVIDIA Jetson Orin Nano provides
8\,GB of on-board memory at 68\,GB/s of bandwidth, whereas a data-center
NVIDIA H100 (SXM5) offers 80\,GB at 3.35\,TB/s---an order of magnitude
difference in capacity and roughly two orders of magnitude in bandwidth.
The batch-size-one regime typical of real-time edge applications further
exacerbates this bottleneck.

We therefore identify the reduction of \emph{active parameter memory},
i.e., the parameters that must be loaded for a single inference, as a key
consideration for efficient inference on memory-constrained edge
accelerators.

\subsubsection{Model Reconstruction for Reduced Memory Footprint}

Existing model compression techniques---pruning, knowledge distillation,
and quantization---reduce the total parameter count
permanently~\cite{han2015deepcompression,hinton2015distillation}. However,
this permanent reduction can degrade the model's representational capacity.

A complementary direction is to exploit input-dependent sparsity in
DNN inference. Empirical evidence indicates that the effective
subset of computation varies with the input: DeepMoE shows that the useful channel subset within a convolutional layer is input-dependent~\cite{wang2019deepmoe}, and SkipNet shows that whole
residual blocks can be skipped per
input~\cite{wang2018skipnet}. Using the full parameter set for every
inference is therefore inherently redundant.

A branched architecture is a natural way to capitalize on this observation
while \emph{decoupling capacity from compute}. Each branch can specialize
to a subset of the input distribution, so the model retains the full dense parameter set rather than permanently removing parameters; meanwhile, at inference, only the backbone plus a single branch is active, directly reducing the memory footprint without
permanently removing parameters.

In this work, we propose \textbf{\boldmath$\Sigma$B} (Sigma-Branch), an umbrella
framework for activation-based hierarchical model reconstruction with
single-path inference. Within this umbrella, we develop a single concrete
instantiation, \textbf{\boldmath$\Sigma$B-Method}, the conversion procedure described
in Section~\ref{sec:method}, and refer to the network produced by it as
\textbf{\boldmath$\Sigma$B-Net}. Throughout the paper, $\Sigma$B denotes the
framework concept, $\Sigma$B-Method denotes the procedure that operates on
a pretrained model, and $\Sigma$B-Net denotes the resulting network whose
inference behavior and empirical numbers are reported. Pretrained models
are restructured into hierarchical branched networks, and only a single
path is executed at inference. The total parameter count is preserved
while the active-parameter memory is substantially reduced, with evaluation
focused on memory-bound edge deployment.

\subsection{Contributions of This Work}\label{sec:intro:contributions}

The contributions of this paper are as follows.

\begin{itemize}
  \item \textbf{Hierarchical Model Reconstruction (C1).} We propose
    $\Sigma$B-Method, a framework that converts pretrained networks into
    hierarchical, MoE-like branched structures, distinguishing our
    approach from flat MoE-style decomposition.
  \item \textbf{Cross-Modal Applicability (C2).} We demonstrate the
    framework on convolutional networks (ResNet-50) and on
    point-cloud networks (PointNet++), establishing applicability across
    two distinct modalities.
  \item \textbf{Extreme Active-Parameter Reduction (C3).} $\Sigma$B-Net
    achieves \textbf{58--60\,\%} active-parameter reduction across
    CIFAR-100/ResNet-50, ImageNet/ResNet-50, and ModelNet40/PointNet++,
    while preserving the classification accuracy of the dense baseline.
    This substantially exceeds state-of-the-art structured pruning
    baselines (FPGM, HRank) at comparable compression ratios.
\end{itemize}

\section{Related Work}\label{sec:related}

We organize previous work into two streams that frame the position of
$\Sigma$B-Method: static structured pruning, which permanently shrinks the
model graph (Section~\ref{sec:related:pruning}), and mixture-of-experts
and hierarchical decomposition, which restructures a dense network to
enable input-dependent computation paths
(Section~\ref{sec:related:moe}). We then position $\Sigma$B-Method against
representative methods along four design axes
(Section~\ref{sec:related:positioning}).

\subsection{Static Structured Pruning}\label{sec:related:pruning}

Static structured pruning permanently removes redundant filters or
channels from a pretrained network in an input-independent manner.
A range of importance criteria has been proposed to identify which
parameters to drop, each offering a distinct view of filter or channel
redundancy and demonstrating measurable compression benefits at
comparable accuracy. Representative importance criteria include geometric median distance among filters
(FPGM~\cite{he2019filter}) and feature-map rank (HRank~\cite{lin2020hrank}).

However, this stream shares a structural limitation that motivates the
present work. Because the parameter removal is permanent, the post-pruning network has strictly lower representational capacity than the dense baseline; this loss has been shown to disproportionately
affect minority-class and atypical samples~\cite{hooker2019selective},
consistent with the representational-capacity motivation discussed in Section~\ref{sec:intro:problem}.

\subsection{Mixture-of-Experts and Hierarchical Decomposition}\label{sec:related:moe}

A complementary stream introduces input-dependent paths by decomposing a
network into experts or a tree of specialized sub-networks. Conventional
mixture-of-experts (MoE) increases capacity by adding many expert branches and routing each input through only the top-$k$ of them:
sparsely-gated MoE~\cite{shazeer2017moe} and
DeepSeekMoE~\cite{dai2024deepseekmoe} both follow this expansion
strategy, in which the total parameter count grows roughly with the
number of experts. While effective for scaling large language models, this is
less suitable for memory-bound edge accelerators, whose on-chip capacity
limits the total, not just the active, parameter count.

A second line preserves the total parameter count of a pretrained dense
network and instead carves it into experts or a tree. DeepMoE introduces
sample-level channel routing into a convolutional backbone by gating
channel subsets with a shallow embedding network~\cite{wang2019deepmoe};
routing is per-input, but the structure remains flat, with no backbone
shared across the per-input sub-networks. A recent line of work analytically
restructures the feed-forward (FFN) sub-layer of a pretrained transformer
into a mixture of experts using neuron activation statistics alone,
requiring no retraining~\cite{pei2026analytical}. The restructuring, however, is confined to FFN sub-layers in transformer-based large language
models: the attention modules remain dense, so the model-wide
active-parameter reduction is bounded by the FFN share of the total
compute rather than applied to the network as a whole, and the
construction does not extend to convolutional or point-cloud backbones.
DecisioNet converts a CNN into a
binary tree of specialized sub-networks and routes each input through a
single path at inference~\cite{gottlieb2023decisionet}, which is
structurally the closest prior work to $\Sigma$B-Method; however, its
tree split is supervised by a class-confusion hierarchy derived from
labels, and its evaluation is confined to convolutional backbones.

\subsection{Positioning of Sigma-Branch Method}\label{sec:related:positioning}

Table~\ref{tab:positioning} positions $\Sigma$B-Method against representative
methods from both streams along four design axes that follow directly
from the requirements established in Section~\ref{sec:intro:problem}.

\begin{table}[t]
  \caption{Positioning of Sigma-Branch Method relative to representative
    compression and dynamic-inference methods. \checkmark{}: satisfied;
    --: not satisfied.}
  \label{tab:positioning}
  \centering
  \footnotesize
  \setlength{\tabcolsep}{4pt}
  \begin{tabular}{l c c c c}
    \toprule
    Method & Hier.\ & Unsup.\ & Sample- & Cross- \\
           & struct.\ & part.\ & level   & modal  \\
    \midrule
    FPGM~\cite{he2019filter}                    & --         & \checkmark & --         & --         \\
    DeepMoE~\cite{wang2019deepmoe}              & --         & \checkmark & \checkmark & --         \\
    Anal.\ FFN-to-MoE~\cite{pei2026analytical}  & --         & \checkmark & --      & --         \\
    DecisioNet~\cite{gottlieb2023decisionet}    & \checkmark & --         & \checkmark & --         \\
    $\Sigma$B-Method (ours)                     & \checkmark & \checkmark & \checkmark & \checkmark \\
    \bottomrule
  \end{tabular}
\end{table}

A \emph{hierarchical} structure enables progressive capacity
decomposition: a shared trunk carries generic features for every sample,
while deep leaves specialize to input clusters, yielding partial sharing that a flat MoE cannot achieve. \emph{Unsupervised} partitioning by activation statistics is required wherever a class-label hierarchy is unavailable or misaligned with the learnable feature structure,
including 3D point-cloud benchmarks for which no canonical label tree
exists. \emph{Sample-level} routing is the natural granularity for image
and point-cloud inputs, for which token-level feed-forward slicing in
transformers does not apply, and it matches the batch-size-one regime of
edge inference. Finally, \emph{cross-modal} validation is required to
substantiate a framework-level claim rather than an architecture-specific
result. To our knowledge, $\Sigma$B-Method is the only method discussed above
that simultaneously satisfies all four axes as required by the
memory-bound edge setting that motivates this work
(Section~\ref{sec:intro:problem}).

\section{Sigma-Branch Method}\label{sec:method}

We now describe $\Sigma$B-Method, a framework that restructures a pretrained
dense network into a hierarchical, single-path inference network we call
$\Sigma$B-Net. The framework consists of four components: a formal
specification of the hierarchical binary-tree architecture
(Section~\ref{sec:method:overview}--\ref{sec:method:formulation}); an
activation-based weight distribution procedure that transfers the pretrained
weights into the new architecture
(Section~\ref{sec:method:actinit}); a soft-routing fine-tuning protocol
with a specialist classification loss and a routing responsibility loss
(Section~\ref{sec:method:finetune}); and a hard top-1 inference procedure
that executes only a single path per input
(Section~\ref{sec:method:infer}). Throughout this section we use a 2-level,
4-leaf instantiation as the canonical example.
Binary routing provides a simple recursive decomposition rule
compatible with the activation-based spherical \(k\)-means
initialization: each split partitions the feature space into two
sub-clusters while keeping the router lightweight and the routing
depth logarithmic in the number of leaves. The \((2,4)\) hierarchy
used throughout this work is therefore intended as a minimal
canonical instantiation rather than a claim of optimal tree size.
The framework generalizes naturally to deeper trees, and we instantiate it on two distinct backbones in Section~\ref{sec:method:crossmodal}.

\subsection{Framework Overview}\label{sec:method:overview}

\begin{figure}[t]
  \centering
  \includegraphics[width=\columnwidth]{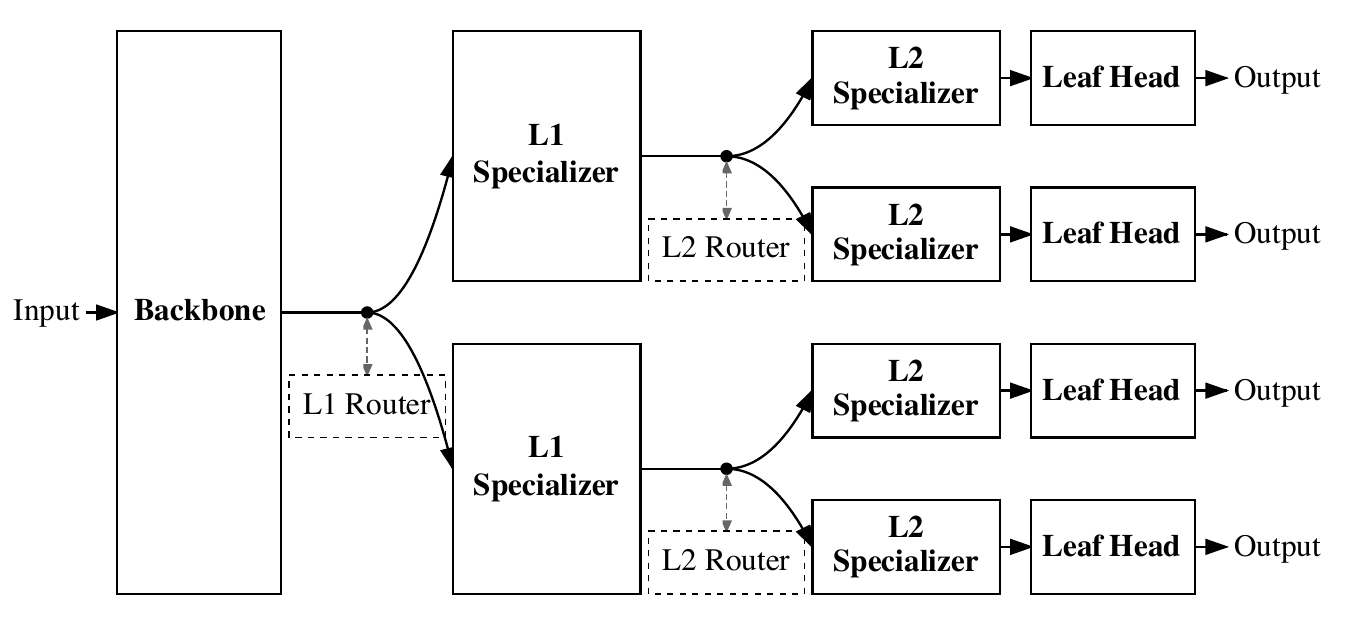}
  \caption{Overall architecture of Sigma-Branch Net in the 2-level, 4-leaf
           canonical configuration. The shared backbone is followed by
           hierarchical binary routers and per-branch specializers, with
           a leaf-specific classification head at every leaf.}
  \label{fig:arch}
\end{figure}

$\Sigma$B-Net is a hierarchical binary tree in which a shared backbone is
followed by routers, branch-specific specializers, and per-leaf
classification heads, as illustrated in Fig.~\ref{fig:arch}. The shared
backbone $f_{\mathrm{BB}}$ processes every input sample and provides the
common feature on which the first router operates. At each level $\ell$,
a router projection $\pi_\ell$ produces a low-dimensional latent, a binary
router $R_\ell$ produces a two-way routing distribution, and two
specializers $s_\ell^{(b)}$ further process the feature conditional on the
chosen branch $b\in\{0,1\}$. At the final level $L$, each leaf $k$ has its
own classification head $\mu_k$. The 2-level, 4-leaf canonical configuration
therefore has one backbone, two level-1 specializers, four level-2
specializers, three routers, and four leaf heads.

Every router is binary, so that the activation-based initialization
of Section~\ref{sec:method:actinit} applies recursively at each
split. The network has two modes of operation: at training time, it
evaluates all leaves and combines them through their routing
probabilities, while at inference time, it selects a single leaf by
hard top-1 routing.

\subsection{Hierarchical Binary-Tree Formulation}\label{sec:method:formulation}

Let $x$ denote an input sample and
$z = f_{\mathrm{BB}}(x)$ the backbone feature. At level $\ell$, the router
projection $u_\ell = \pi_\ell(\cdot)$ consists of an average pool, a linear
layer, layer normalization, and dropout, and produces a low-dimensional
latent. The binary router applies a linear map followed by softmax,
\begin{equation}
  p^{(\ell)} = \mathrm{softmax}\!\bigl(R_\ell(u_\ell)\bigr) \in \Delta^{1},
  \label{eq:router}
\end{equation}
where $\Delta^{1}$ denotes the 1-simplex. The probability of reaching
leaf $k$ factorizes hierarchically as the product of per-level
probabilities,
\begin{equation}
  p_{\mathrm{leaf}}(k \mid x) = \prod_{\ell=1}^{L} p^{(\ell)}_{b_\ell(k)},
  \label{eq:pleaf}
\end{equation}
where $b_\ell(k)\in\{0,1\}$ denotes the branch chosen at level $\ell$ on
the path to leaf $k$. The output of leaf $k$ is obtained by composing the
specializers along that path and then applying the leaf head,
\begin{equation}
  y_k = \mu_k\!\bigl(s_L^{(b_L(k))} \circ \cdots \circ s_1^{(b_1(k))}(z)\bigr).
  \label{eq:leaf-out}
\end{equation}
At training time, the network output is the routing-weighted combination
$\hat{y}_{\mathrm{soft}} = \sum_{k=1}^{K} p_{\mathrm{leaf}}(k \mid x)\, y_k$;
at inference, only the single leaf with the highest routing probability is evaluated,
as detailed in Section~\ref{sec:method:infer}.

The number of parameters that must be loaded at inference,
\begin{equation}
  P_{\mathrm{active}}
    = |\theta_{\mathrm{BB}}|
      + \sum_{\ell=1}^{L} |\theta_{s_\ell^{(b_\ell(k^*))}}|
      + |\theta_{\mu_{k^*}}|,
  \label{eq:active}
\end{equation}
is therefore proportional to a single root-to-leaf path, while the
total parameter count
$P_{\mathrm{total}} = |\theta_{\mathrm{BB}}| + 2|\theta_{s_1}|
+ 4|\theta_{s_2}| + 4|\theta_{\mu}|$ in the 4-leaf configuration is
substantially larger. To prevent $P_{\mathrm{total}}$ from growing
beyond the baseline as $n$ parallel branches are introduced at a level, we
reduce the channel width of each specializer by a factor of $1/\sqrt{n}$
relative to the baseline block, so that the overall parameter budget remains comparable to the baseline. In our canonical ResNet-50 instantiation, the resulting
widths are $512$ for the shared backbone, $\approx\!512\sqrt{2}\approx 724$
for each level-1 specializer, and $\approx\!724\sqrt{2}\approx 1024$ for
each level-2 specializer.

\subsection{Activation-Based Weight Distribution}\label{sec:method:actinit}

Jointly training a router and a set of experts from scratch in
MoE-style architectures is known to suffer from a chicken-and-egg
dependency, as observed in the original sparsely-gated formulation by
Shazeer et al.~\cite{shazeer2017moe}. In our setting the same
dependency arises between the router and the specializers: the router
cannot send samples to the appropriate branches unless the specializers are already specialized for distinct input subsets, while the
specializers cannot specialize unless the router consistently feeds
them their respective subsets. Without an external intervention, soft routing combined with all-leaf training resolves this circularity in
the trivial direction in which all branches converge toward similar
functions, and branch specialization never emerges.

We therefore initialize the router and the specializers
\emph{jointly}, from a single source of structure, rather than
through independent procedures.
We eliminate this mismatch by construction: a single spherical
$k$-means clustering of the projection latents provides both the
router weight, as cluster centroids, and the per-branch channel
allocation, via cluster-conditional source-layer activations, so the router's per-sample decision and the specializer's per-cluster channel inventory are aligned from the outset.

\textit{Phase~0: backbone transfer.} The early generic-feature layers of
the pretrained baseline are copied verbatim into $f_{\mathrm{BB}}$. Their
shapes are designed to match, so the transfer is a direct parameter copy
and preserves the baseline's low-level feature extractor.

\textit{Phase~1: level-1 clustering and router initialization.} We forward
the training set through $f_{\mathrm{BB}}$ and the level-1 projection
$\pi_1$, collecting up to $N$ latents
$\{u_1^{(i)}\}_{i=1}^{N}$ with $N \le 50{,}000$. We then run a spherical
$k$-means algorithm with $k=2$~\cite{dhillon2001concept,banerjee2005clustering}
on these latents to obtain centroids $C_1 \in \mathbb{R}^{2\times D}$ and
cluster assignments $c_i \in \{0,1\}$. Setting $R_1.W \leftarrow C_1$ and
$R_1.b \leftarrow 0$ makes
$\mathrm{softmax}\!\bigl(C_1\, u_1\bigr)$ a cosine-similarity-based binary
router whose decision boundary coincides with the $k$-means partition. In
parallel, we forward the same samples through the baseline up to the
level-1 source block and accumulate the pooled per-channel mean activation
$a^{(c)}$ conditioned on each cluster $c$.

\begin{figure}[t]
  \centering
  \includegraphics[width=\columnwidth]{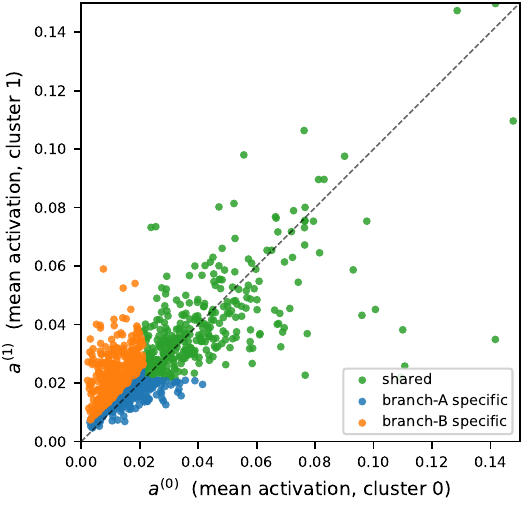}
  \caption{Shared and branch-specific channel selection on the level-1
           source block of a pretrained ResNet-50 over ImageNet-1k.
           Each point is one of the $1024$ output channels of the
           source layer, plotted by its cluster-conditional mean
           activations $a^{(0)}$ and $a^{(1)}$. Channels with large
           $\min(a^{(0)}, a^{(1)})$, i.e.\ those strongly activated in
           both clusters, are selected as shared; the remaining
           channels with large $|a^{(0)} - a^{(1)}|$ are selected as
           branch-specific. The axes are clipped at $0.15$ so that
           the bulk of the channel distribution is visible; a small
           number of high-activation outliers in the shared set fall
           outside the panel.}
  \label{fig:chansel}
\end{figure}

\textit{Phase~2: shared and branch-specific channel selection.} The
level-1 specializer width is $W_{L1}$; we allocate
$K_{\mathrm{shared}} = \lfloor \kappa W_{L1} \rfloor$ channels as shared
between the two branches and $W_{L1} - K_{\mathrm{shared}}$ as
branch-specific, where $\kappa \in [0,1]$ controls the trade-off and is
fixed to $0.5$ in our experiments. The shared channels are chosen as the
top-$K_{\mathrm{shared}}$ indices of
$\min\!\bigl(a^{(0)}, a^{(1)}\bigr)$, so that a channel qualifies as
shared only when it is strongly activated in \emph{both} clusters; this
prevents channels that are strong in only one cluster from being routed
to the shared pool. The branch-specific channels for cluster $c$ are
then chosen from the remaining channels by the contrast score
$a^{(c)} - a^{(1-c)}$, capturing features that are relatively more
active for that cluster. For convolutional baselines, the bottleneck's internal
plane indices are additionally selected by L1-norm and shared across the
two branches. The resulting index set defines which baseline channels are
copied into $s_1^{(A)}$ and $s_1^{(B)}$. Fig.~\ref{fig:chansel} visualizes the partition for our ResNet-50 instantiation: each baseline channel is
plotted as one point in the $(a^{(0)}, a^{(1)})$ plane, with channels near
the diagonal becoming shared and the off-diagonal channels becoming
branch-specific. The decomposition mirrors the shared-expert pattern
of DeepSeekMoE~\cite{dai2024deepseekmoe}, where always-active shared
experts free routed experts to specialize sharply; in our setting,
shared channels carry universally useful features while
branch-specific channels capture cluster-distinctive ones.

\textit{Phases~3--4: level-2 recursion.} Using the initialized
level-1 components, we forward each sample only through the branch chosen
by $R_1$, collecting per-branch latents $u_A$ and $u_B$ at the level-2
projection. Spherical $k$-means with $k=2$ is then applied within each
branch to obtain centroids $C_A,C_B$, which initialize $R_{L2,A}$ and
$R_{L2,B}$. Cluster-conditional activations of the level-2 source block
are aggregated within each branch, and the shared and branch-specific
selection of Phase~2 is repeated to produce the four level-2 specializers
$s_2^{(A1)}, s_2^{(A2)}, s_2^{(B1)}, s_2^{(B2)}$.

\textit{Phase~5: leaf head transfer.} Finally, each leaf head $\mu_k$
inherits the column subset of the baseline classifier weight that
corresponds to its allocated output channels. The classifier bias,
which is indexed by output class and does not depend on the input
channels, is shared identically across all leaf heads. Each leaf head
thus starts from a partial copy of the baseline classifier restricted
to the channel subset that its specializer produces, and the four
leaves differ only in the columns of the inherited weight matrix.

Geometrically, the procedure performs activation-based hierarchical
partitioning of the input distribution: each level partitions the
remaining feature space by spherical $k$-means and inherits the baseline
channels that are most active on the resulting clusters. The router
projection $\pi_\ell$ is kept close to linear during initialization by
omitting the inner non-linearity, so that the spherical structure of the
$k$-means cluster centroids in projection space carries over directly to the router weights.

\begin{algorithm}[t]
\caption{Activation-based weight distribution for Sigma-Branch Net}
\label{alg:actinit}
\begin{algorithmic}[1]
\Require Pretrained baseline $\mathcal{M}_{\mathrm{base}}$; training set
         $\mathcal{D}$; shared-channel ratio $\kappa$; depth $L$
\Ensure Initialized $\Sigma$B-Net parameters
\State $f_{\mathrm{BB}}.\theta \gets$ shared-trunk weights of
       $\mathcal{M}_{\mathrm{base}}$ \Comment{Phase 0}
\For{$\ell = 1$ \textbf{to} $L$}
    \State Collect latents
           $\{u_\ell^{(i)} = \pi_\ell(z_\ell^{(i)})\}_{i=1}^{N}$ over
           $\mathcal{D}$
    \State $(C_\ell,\, c_i)
            \gets \textsc{SphericalKMeans}(\{u_\ell^{(i)}\},\, k=2)$
           \Comment{Phase $\ell$-a}
    \State $R_\ell.W \gets C_\ell,\; R_\ell.b \gets 0$
    \For{branch $b \in \{0,1\}$ at level $\ell$}
        \State $a^{(c)} \gets$ cluster-conditional pooled activation of
               baseline level-$\ell$ source layer
        \State $\mathrm{shared} \gets
               \arg\mathrm{top}_{\lfloor \kappa W_\ell \rfloor}\!\bigl(\min(a^{(0)}, a^{(1)})\bigr)$
        \State $\mathrm{specific}_b \gets
               \arg\mathrm{top}_{W_\ell - \lfloor \kappa W_\ell \rfloor}\!\bigl(a^{(b)} - a^{(1-b)}\bigr)$
        \State $\mathrm{out\_idx}_b
               \gets \mathrm{shared} \cup \mathrm{specific}_b$
        \State Copy baseline level-$\ell$ block channels indexed by
               $\mathrm{out\_idx}_b$ into $s_\ell^{(b)}$
               \Comment{Phase $\ell$-b}
    \EndFor
\EndFor
\For{each leaf $k$}
    \State $\mu_k.\mathrm{fc} \gets
           \mathcal{M}_{\mathrm{base}}.\mathrm{fc}[:,\, \mathrm{out\_idx}_k]$
           \Comment{Phase $L{+}1$}
\EndFor
\end{algorithmic}
\end{algorithm}

\subsection{Soft-Routing Fine-Tuning}\label{sec:method:finetune}

After activation-based initialization, the backbone, routers, specializers, and leaf heads are all initialized from the pretrained
baseline; only the router projections $\pi_\ell$ have no baseline
counterpart and remain randomly initialized. The router projections are
nevertheless used during the spherical $k$-means step of Phases~1 and~3, and the resulting cluster structure is preserved because spherical $k$-means depends only on angular similarity, which random linear projections approximately preserve in expectation~\cite{johnson1984extensions}.
After initialization, all parameters are jointly fine-tuned with soft routing, in which every leaf is evaluated for every input, and the losses are combined through the routing probabilities.

We use a \emph{specialist} classification loss in which the leaf-wise cross-entropy losses are weighted by the routing probabilities
$p_{\mathrm{leaf}}$:
\begin{equation}
  \mathcal{L}_{\mathrm{cls}}
    = \mathbb{E}_x\!\left[\sum_{k=1}^{K}
      p_{\mathrm{leaf}}(k \mid x)\,
      \mathrm{CE}\!\bigl(y_k(x),\, y\bigr)\right].
  \label{eq:loss-cls}
\end{equation}
This formulation differs from a naive ensemble cross-entropy on the
weighted-average logits in that each leaf must itself produce a correct
prediction when its routing probability is high, which directly aligns the soft-routing training objective with the hard top-1 inference behavior at test time.

To further align the router decision with the per-leaf prediction
quality, we augment the objective with a responsibility-matching loss
adapted from the gating literature. The responsibility $r_k$ of leaf $k$
is defined by the soft minimum of the leaf-wise cross-entropy losses,
\begin{equation}
  r_k(x)
    = \frac{\exp\!\bigl(-\mathrm{CE}(y_k(x), y)/\tau_r\bigr)}
           {\sum_{j}\exp\!\bigl(-\mathrm{CE}(y_j(x), y)/\tau_r\bigr)},
  \label{eq:resp}
\end{equation}
and is treated as a stop-gradient target. The routing probabilities are
trained to match this responsibility distribution through a cross-entropy
penalty,
\begin{equation}
  \mathcal{L}_{\mathrm{resp}}
    = -\,\mathbb{E}_x\!\left[\sum_{k} r_k(x)\,
      \log p_{\mathrm{leaf}}(k \mid x)\right].
  \label{eq:loss-resp}
\end{equation}
The total training objective is
\begin{equation}
  \mathcal{L}
    = \mathcal{L}_{\mathrm{cls}}
      + \lambda_{\mathrm{resp}}\, \mathcal{L}_{\mathrm{resp}},
  \label{eq:loss-total}
\end{equation}
where $\lambda_{\mathrm{resp}}$ is a fixed scalar coefficient and
$\tau_r$ is a fixed scalar that controls the sharpness of the
responsibility distribution. We set them to $\lambda_{\mathrm{resp}} = 0.3$ and $\tau_r = 0.3$.

\subsection{Hard Top-1 Inference}\label{sec:method:infer}

At inference, $\Sigma$B-Net evaluates only a single root-to-leaf path.
Starting from the backbone feature $z = f_{\mathrm{BB}}(x)$, the level-1 router selects a branch by hard argmax,
\begin{equation}
  b_\ell^{*} = \arg\max_{j\in\{0,1\}}\, p^{(\ell)}_{j},
  \label{eq:argmax}
\end{equation}
and only the chosen specializer $s_\ell^{(b_\ell^{*})}$ is executed.
The selection is sequential rather than joint: the level-2 projection input is the output of the chosen level-1 specializer, so the level-2 routing distribution exists only along the branch chosen at level~1. Once the leaf $k^{*}$ corresponding to the sequence of argmax decisions is determined, the prediction is taken directly from the
selected leaf head, $\hat{y} = y_{k^{*}}$.

The active-parameter count of one inference is therefore
$P_{\mathrm{active}}$ from \eqref{eq:active}, comprising the backbone, one specializer per level, and one leaf head. In the 4-leaf configuration, three of the four leaf heads, three of the four level-2 specializers, and one of the two level-1 specializers are entirely
skipped.

\subsection{Modality-Agnostic Applicability}\label{sec:method:crossmodal}

The only assumption the framework places on the baseline is that
$f_{\mathrm{BB}}$ produces a pooled global feature on which the router projection $\pi_\ell$ can operate; the spherical $k$-means clustering, the cluster-conditional channel selection, and the per-leaf path selection are otherwise modality-independent. The framework therefore extends to any such backbone, including convolutional networks (global average pooling) and PointNet++-style networks (set-abstraction descriptors). Per-modality choices of source block, specializer width, and training schedule are reported in Section~\ref{sec:setup}.

\section{Experimental Setup}\label{sec:setup}

\subsection{Hardware and Software Environment}\label{sec:setup:hw}

All models are trained with
NVIDIA A100 GPUs. The implementation uses PyTorch with the timm
training utilities.

\subsection{Models and Datasets}\label{sec:setup:data}

We instantiate \(\Sigma\)B-Method on three backbone--dataset combinations, covering image classification (ResNet-50 on CIFAR-100,  ResNet-50 on ImageNet-1K) and point-cloud classification (PointNet++ on ModelNet40).

\subsubsection{CIFAR-100 with ResNet-50}
CIFAR-100~\cite{krizhevsky2009learning} contains 50{,}000 training and
10{,}000 test images at $32{\times}32$ resolution across 100 classes.
The baseline is a timm ResNet-50~\cite{he2016resnet} with the CIFAR-style modification of a $3{\times}3$ stride-1 stem and no max-pool. The \(\Sigma\)B-Net counterpart uses the stem and layers~1--2 as the shared trunk (backbone width 512), six bottleneck blocks per level-1 specializer at width 724 (\(\approx 512\sqrt{2}\)), and three blocks per level-2 specializer at width 1024 (\(\approx 724\sqrt{2}\)).
All four leaves share the same width and depth, so 
\(P_{\mathrm{active}}\) is path-invariant. The projection dimension
used by the spherical $k$-means router is 128.

\subsubsection{ImageNet-1K with ResNet-50}
ImageNet-1K~\cite{deng2009imagenet} consists of 1{,}281{,}167 training
and 50{,}000 validation images across 1{,}000 classes. The baseline is
a timm ResNet-50 with the standard $7{\times}7$ stride-2 stem followed
by max-pool. The \(\Sigma\)B-Net uses the same specializer widths as
the CIFAR-100 instantiation; only the stem is replaced with the
ImageNet variant.

\subsubsection{ModelNet40 with PointNet++}
ModelNet40~\cite{wu20153d} consists of 9{,}843 training and
2{,}468 test point clouds across 40 classes, uniformly sampled to
1{,}024 points per shape. The baseline is the
PointNet++~\cite{qi2017pointnetpp} single-scale-grouping variant. The
\(\Sigma\)B-Net uses SA1 and SA2 as the shared trunk (global feature
dimension 256), an SA3-equivalent ball-query MLP at width 512 for each
level-1 specializer, an FC1-equivalent MLP at width 256 for each
level-2 specializer, and an FC2-equivalent head per leaf. 

\subsection{Training Protocol}\label{sec:setup:train}

Table~\ref{tab:train} summarizes the training recipe for each
combination. For image classification, we follow the timm-style training
protocol: SGD with Nesterov momentum, an initial learning rate of
\(0.1\), cosine annealing~\cite{loshchilov2017sgdr} with five warm-up
epochs, and label smoothing of \(0.1\). CIFAR-100 runs for 200 epochs
with weight decay \(10^{-4}\); ImageNet-1K runs for 90 epochs. For point-cloud
classification we follow the PointNet++ recipe: Adam~\cite{kingma2017adam}
with initial learning rate \(10^{-3}\), weight decay \(10^{-4}\), and
step decay (\(\gamma = 0.7\) every 20 epochs) for 200 epochs.

\begin{table*}[!t]
  \centering
  \caption{Training hyper-parameters per backbone--dataset combination.}
  \label{tab:train}
  \small
  \begin{tabular}{lccc}
    \toprule
    Setting & CIFAR-100 / ResNet-50 & ImageNet-1K / ResNet-50 & ModelNet40 / PointNet++ \\
    \midrule
    Optimizer              & SGD (Nesterov) & SGD (Nesterov) & Adam \\
    Initial learning rate  & \(0.1\) & \(0.1\) & \(10^{-3}\) \\
    Weight decay           & \(10^{-4}\) & \(10^{-4}\) & \(10^{-4}\) \\
    LR schedule            & cosine, 5\,ep.\ warm-up & cosine, 5\,ep.\ warm-up & step (\(\gamma=0.7\) per 20\,ep.) \\
    Batch size             & 256 & 256 & 32 \\
    Epochs                 & 200 & 90 & 200 \\
    Label smoothing        & \(0.1\) & \(0.1\) & \(0.0\) \\
    \bottomrule
  \end{tabular}
\end{table*}

\subsection{Baselines for Comparison}\label{sec:setup:baselines}

We organize the comparison around three baseline families, each chosen
to isolate a specific axis of the trade-off space rather than to claim
head-to-head dominance on every axis.

\subsubsection{Original models (D-1)}
The first comparison is against the unconverted backbones themselves:
ResNet-50~\cite{he2016resnet} on CIFAR-100 and ImageNet-1K, and
PointNet++~\cite{qi2017pointnetpp} on ModelNet40. Each baseline is
retrained in our environment under the protocol of
Table~\ref{tab:train}, and we report accuracy, FLOPs, and active
parameters before and after \(\Sigma\)B-Method is applied. This
establishes whether the reconstructed network preserves the predictive
quality of its source model.

\subsubsection{Static filter pruning (D-2)}
We compare against two representative static filter-pruning methods on
ResNet-50: FPGM~\cite{he2019filter} and HRank~\cite{lin2020hrank}, both
of which report ImageNet-1K ResNet-50 numbers at multiple compression
ratios from their own dense baseline. This baseline
characterizes the static FLOPs-reduction frontier against which the dynamic-path approach is positioned.

\subsubsection{Sample-level dynamic computation (D-3)}
We compare against SkipNet~\cite{wang2018skipnet}, which augments a
ResNet with recurrent gating modules that decide, per input, whether
each residual block is executed or skipped, and against the narrow
variant of DeepMoE~\cite{wang2019deepmoe}, which gates individual
output channels of each convolution via a shallow embedding network.
Both methods share with \(\Sigma\)B-Net the property of
input-conditional path selection, but differ in granularity:
SkipNet operates at the block level with stochastic binary gates,
DeepMoE at the channel level with continuous ReLU gates, and
\(\Sigma\)B-Net at the leaf level with hard top-1 routing through a
hierarchical tree.

Two further entries appearing in the positioning table
(Table~\ref{tab:positioning})---DecisioNet~\cite{gottlieb2023decisionet}
and the analytical FFN-to-MoE conversion
of~\cite{pei2026analytical}---are retained as related work, but
excluded from numerical comparison. DecisioNet builds a hierarchical
binary tree of specialized paths, but trains the network from scratch
using a \emph{supervised} label-derived class hierarchy; the \(\Sigma\)B-Method, by contrast, fine-tunes a pretrained dense model using an \emph{unsupervised} activation-based partition
(Section~\ref{sec:method:actinit}). A head-to-head comparison would, therefore, conflate the training regime with the architectural design.
The analytical FFN-to-MoE method targets the feed-forward sub-layers
of a pretrained Transformer with no retraining, and does not apply
to convolutional or point-cloud backbones, which are within the scope of
our experiments. Both methods remain in the related-work positioning
(Section~\ref{sec:related}) to clarify where
\(\Sigma\)B-Method sits in the design space.

\subsection{Evaluation Metrics}\label{sec:setup:metrics}

For each method and combination, we report the following metrics.

(i)~\textbf{Top-1 and Top-5 accuracy} on the held-out test set, or the
validation set for ImageNet-1K.

(ii)~\textbf{Active parameters} \(P_{\mathrm{active}}\) as defined in
Equation~\eqref{eq:active}, counted along the hard-routed path.
The active-parameter metric is intended as an analytical proxy
for per-inference off-chip weight-transfer volume under
memory-bound deployment, rather than as a direct measurement
of wall-clock latency.

(iii)~\textbf{FLOPs}, measured with fvcore on a forward pass under
hard top-1 routing with batch size one, so that the data-dependent
routing short-circuits to a single executed path.

(iv)~\textbf{Routing distribution}, a qualitative analysis of which
leaves are selected for each class. The per-class leaf usage is visualized as a heatmap and serves as an interpretability check on
the unsupervised partition rather than a quantitative score.

Across~(i)--(iii), the three baseline families differ in how they
occupy the FLOPs--\(P_{\mathrm{active}}\) plane: static pruning
compresses both axes by similar fractions but does so permanently,
sample-level dynamic computation reduces both axes through input-
dependent soft gating, and \(\Sigma\)B-Method reduces both axes
through input-dependent hard top-1 routing while retaining the full dense parameter set. Section~\ref{sec:results} reports the
corresponding numerical comparison.

\section{Results}\label{sec:results}

\subsection{Accuracy and Compute Trade-off across Tasks}\label{sec:results:main}

Table~\ref{tab:results_main} summarizes the main results across the
three backbone--dataset combinations. The primary observation is that
\(\Sigma\)B-Net reduces the active-parameter footprint
\(P_{\mathrm{active}}\) by 58--60\,\% relative to its dense baseline
while remaining either within 0.1\,pp of the baseline accuracy
(CIFAR-100), within 1.7\,pp (ImageNet-1K), or above it
(ModelNet40). FLOPs at hard top-1 inference are reduced by 10--32\,\%, with the variation reflecting how much of the network is shared across all inputs in each backbone.

\begin{table*}[!t]
  \centering
  \caption{Main results across the three backbone--dataset
  combinations. Active param.~\(\downarrow\) is the reduction in
  \(P_{\mathrm{active}}\) relative to the dense baseline.}
  \label{tab:results_main}
  \footnotesize
  \setlength{\tabcolsep}{4pt}
  \begin{tabular}{llcccccccc}
    \toprule
    Dataset & Backbone & Method & Epochs & Top-1 (\%) & Top-5 (\%) & \(P_{\mathrm{total}}\) (M) & \(P_{\mathrm{active}}\) (M) & Active param.~\(\downarrow\) (\%) & FLOPs (G) \\
    \midrule
    CIFAR-100    & ResNet-50  & Baseline       & 200       & 76.99      & 94.65 & 23.71 & 23.71 & ---  & 1.305 \\
    CIFAR-100    & ResNet-50  & \(\Sigma\)B-Net & 60 (FT)  & 76.92      & 95.30 & 25.55 & \textbf{9.41}  & 60.3 & 0.888 \\
    ImageNet-1K  & ResNet-50  & Baseline       & 90        & 76.54      & --- & 25.56 & 25.56 & ---  & 4.112 \\
    ImageNet-1K  & ResNet-50  & \(\Sigma\)B-Net & 100 (FT) & 74.82      & 91.58 & 29.25 & \textbf{10.34} & 59.5 & 2.834 \\
    ModelNet40   & PointNet++ & Baseline       & 200       & 90.15      & 99.19 & 1.47  & 1.47  & ---  & 1.148 \\
    ModelNet40   & PointNet++ & \(\Sigma\)B-Net & 200 (FT) & \textbf{91.25}      & 99.11 & 1.46  & \textbf{0.61}  & 58.3 & 1.030 \\
    \bottomrule
  \end{tabular}
\end{table*}

On CIFAR-100 with ResNet-50, the \(\Sigma\)B-Net obtained from a
60-epoch fine-tune of the 200-epoch dense baseline retains Top-1
within 0.07\,pp (76.99\,\% \(\to\) 76.92\,\%) while reducing
\(P_{\mathrm{active}}\) by 60.3\,\%. Top-5 also improves by
0.65\,pp. Hard-routed FLOPs drop by 32.0\,\%.

On ImageNet-1K with ResNet-50, the \(\Sigma\)B-Net is fine-tuned for
100 epochs from a 90-epoch dense baseline. Top-1 drops by 1.72\,pp
(76.54\,\% \(\to\) 74.82\,\%) while \(P_{\mathrm{active}}\) falls
by 59.5\,\% and FLOPs by 31.1\,\%.

On ModelNet40 with PointNet++, the \(\Sigma\)B-Net improves Top-1
by 1.10\,pp (90.15\,\% \(\to\) 91.25\,\%) while reducing
\(P_{\mathrm{active}}\) by 58.3\,\%. FLOPs reduction is modest
(10.3\,\%) because the shared trunk SA1 + SA2, which is executed for
every input, dominates the per-sample compute on point clouds.

\subsection{Comparison with Static Structured Pruning}\label{sec:results:static}

Table~\ref{tab:results_static} compares \(\Sigma\)B-Net on
ImageNet-1K ResNet-50 against two representative static
filter-pruning methods that report results on their own dense baseline:
FPGM~\cite{he2019filter} at two pruning rates, and
HRank~\cite{lin2020hrank} at three sparsity levels. We use the
numbers reported in the original publications.

\begin{table*}[!t]
  \centering
  \caption{Comparison with static structured pruning on ImageNet-1K
  ResNet-50. \(\S\) marks analytic values derived from the published
  FPGM pruning rule.}
  \label{tab:results_static}
  \footnotesize
  \setlength{\tabcolsep}{4pt}
  \begin{tabular}{lcccccc}
    \toprule
    Method & Top-1 (\%) & Top-5 (\%) & Active param. (M) & FLOPs (G) & FLOPs\(\downarrow\) (\%) & Active param.~\(\downarrow\) (\%) \\
    \midrule
    ResNet-50 baseline~\cite{lin2020hrank}    & 76.15 & 92.87 & 25.50           & 4.09 & ---  & --- \\
    ResNet-50 baseline (ours)                  & 76.54 & ---   & 25.56           & 4.11 & ---  & --- \\
    FPGM-30~\cite{he2019filter}                & 75.59 & 92.63 & 16.96\(^{\S}\)  & 2.36\(^{\S}\) & 42.2 & 33.6\(^{\S}\) \\
    FPGM-40~\cite{he2019filter}                & 74.83 & 92.32 & 14.57\(^{\S}\)  & 1.90\(^{\S}\) & 53.5 & 43.0\(^{\S}\) \\
    HRank-44\%~\cite{lin2020hrank}             & 74.98 & 92.33 & 16.15           & 2.30 & 43.8 & 36.7 \\
    HRank-62\%~\cite{lin2020hrank}             & 71.98 & 91.01 & 13.77           & 1.55 & 62.1 & 46.0 \\
    HRank-76\%~\cite{lin2020hrank}             & 69.10 & 89.58 & 8.27            & 0.98 & 76.0 & 67.6 \\
    \(\Sigma\)B-Net (ours)                      & \textbf{74.82} & \textbf{91.58} & \textbf{10.34} & \textbf{2.834} & 31.1 & \textbf{59.5} \\
    \bottomrule
  \end{tabular}
\end{table*}

At a comparable Top-1 (74.8--75.0\,\%), \(\Sigma\)B-Net's
\(P_{\mathrm{active}}\) reduction is 14--23\,pp larger than that of the static methods: 59.5\,\% versus 43.0\,\% for FPGM-40 (Top-1
74.83\,\%) and 36.7\,\% for HRank-44 (Top-1 74.98\,\%). On the FLOPs axis the trade-off is reversed: at the same Top-1 band, the static methods remove 42.2--53.5\,\% of FLOPs while \(\Sigma\)B-Net removes
31.1\,\%. The implications of this asymmetric position on the
active-parameter--FLOPs plane are discussed in
Section~\ref{sec:discussion:position}.

\subsection{Comparison with Sample-Level Dynamic Computation}\label{sec:results:dynamic}

We compare \(\Sigma\)B-Net against two representative
sample-level dynamic-inference methods on ImageNet-1K ResNet-50:
SkipNet~\cite{wang2018skipnet}, which selects per-input residual
blocks to skip, and the narrow variant of
DeepMoE~\cite{wang2019deepmoe}, which gates output channels per input via a shallow embedding network.
Table~\ref{tab:results_dynamic} reports the comparison; active-parameter and
FLOPs reductions are computed per input from gate masks on the
ImageNet-1K validation set and averaged.

\begin{table}[!t]
  \centering
  \caption{Comparison with sample-level dynamic computation on
  ImageNet-1K ResNet-50.}
  \label{tab:results_dynamic}
  \footnotesize
  \setlength{\tabcolsep}{4pt}
  \begin{tabular}{lccc}
    \toprule
    Method & Top-1 (\%) & Active param.~\(\downarrow\) (\%) & FLOPs\(\downarrow\) (\%) \\
    \midrule
    \(\Sigma\)B-Net (ours)                   & \textbf{74.82} & \textbf{59.5} & 31.1 \\
    SkipNet+SP~\cite{wang2018skipnet}        & 70.03          & 6.7           & 6.7  \\
    Narrow-DeepMoE-A~\cite{wang2019deepmoe}  & 66.27          & 43.6          & 59.6 \\
    \bottomrule
  \end{tabular}
\end{table}

\(\Sigma\)B-Net reports both the highest Top-1 and the largest
active-parameter reduction in
Table~\ref{tab:results_dynamic}. The structural distinction is
that each input selects exactly one root-to-leaf path of
\(\Sigma\)B-Net, fixing the active-parameter mass by architecture
rather than by trained gate values; SkipNet's per-block Bernoulli
gating and DeepMoE's continuous channel gating both produce
sample-dependent active footprints, which complicate deterministic
resource provisioning on edge accelerators.

\subsection{Routing Behavior}\label{sec:results:routing}

We inspect the learned routing on the CIFAR-100 \(\Sigma\)B-Net. The
validation-set leaf utilization under hard top-1 routing is
\((A_1, A_2, B_1, B_2) = (20.0, 23.6, 13.9, 42.6)\,\%\), giving a
balance score of 0.935 on the \([0, 1]\) scale where one denotes
the uniform distribution and zero denotes complete collapse to a
single leaf. The routers are well separated, and no leaf is starved.

\begin{figure}[!t]
  \centering
  \includegraphics[width=0.85\columnwidth]{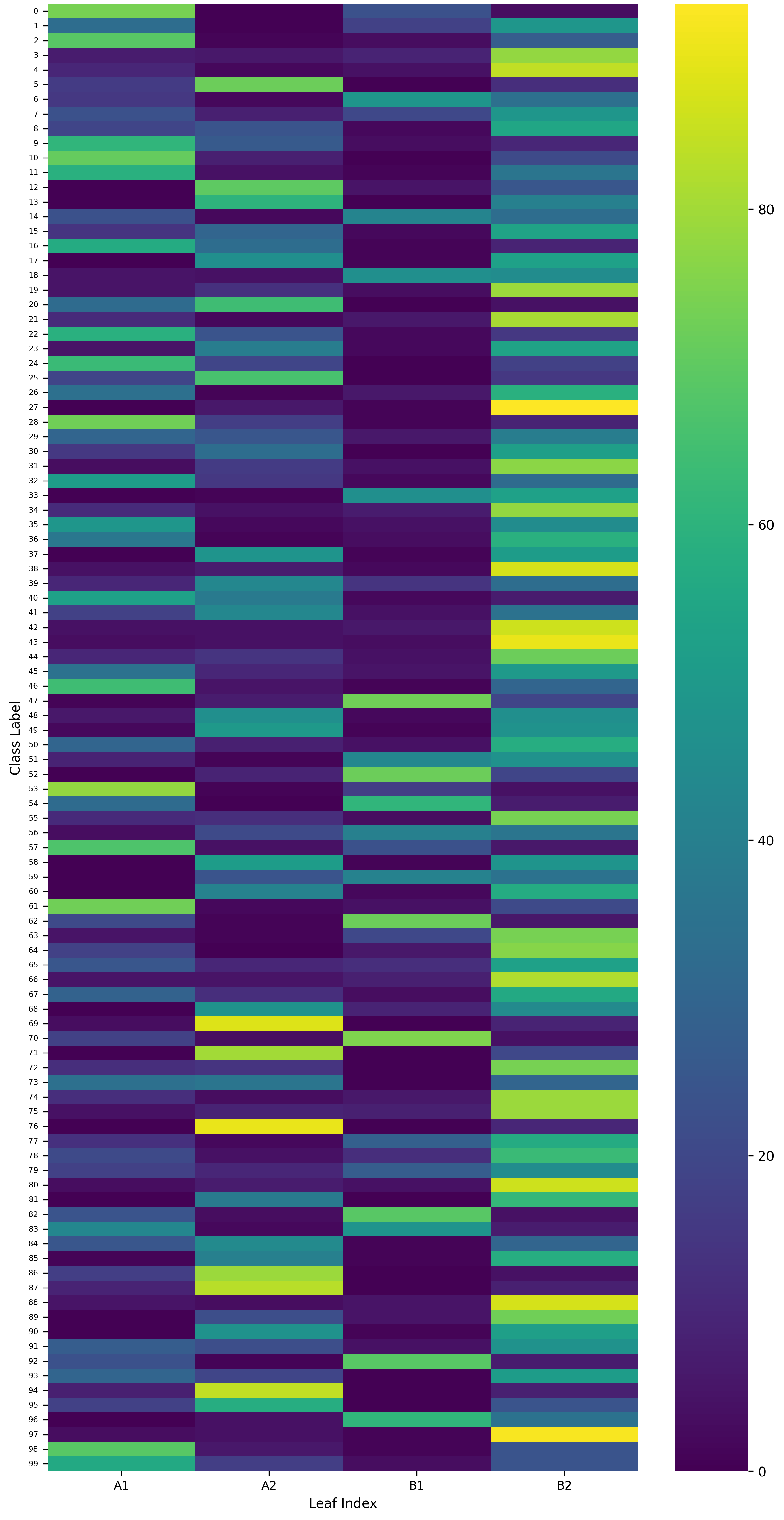}
  \caption{Per-class leaf-assignment heatmap for Sigma-Branch Net on
  CIFAR-100 (seed-42 reproduction of the main-table 
  \(\Sigma\)B-Net configuration). Rows are the 100 CIFAR-100 classes
  (re-ordered by dominant leaf); columns are the four leaves
  \((A_1, A_2, B_1, B_2)\); cell intensity is the fraction of
  validation samples of that class routed to that leaf under hard
  top-1 routing.}
  \label{fig:routing_heatmap}
\end{figure}

Fig.~\ref{fig:routing_heatmap} shows the per-class leaf-assignment
distribution: for each of the 100 CIFAR-100 classes, we report the
fraction of validation samples routed to each leaf under hard top-1.
The heatmap exhibits a clear block structure: distinct subsets of classes concentrate on different leaves, and most classes route into a single dominant leaf rather than distributing uniformly. This is the qualitative signature predicted by the activation-based initialization in Section~\ref{sec:method:actinit}: by construction, the leaves are seeded from cluster-conditional channel subsets, and the responsibility loss in Section~\ref{sec:method:finetune} preserves this clustering through fine-tuning.

\subsection{Ablations}\label{sec:results:ablation}

We ablate the two non-trivial design choices in
Section~\ref{sec:method}: the activation-based channel partition used
by the pretrained-baseline initialization
(Section~\ref{sec:method:actinit}) and the routing-responsibility
loss (Section~\ref{sec:method:finetune}). All ablations are run on
ImageNet-1K with ResNet-50 under the same fine-tuning recipe as the
main result. The default configuration matches the main-results row.

\begin{table*}[!t]
  \centering
  \caption{Ablations on ImageNet-1K ResNet-50. Balance is the
  entropy-normalized leaf utilization (one = uniform).}
  \label{tab:results_ablation}
  \footnotesize
  \setlength{\tabcolsep}{6pt}
  \begin{tabular}{lccccc}
    \toprule
    Variant & Init mode & \(\lambda_{\mathrm{resp}}\) & Top-1 (\%) & Balance & Leaf usage \% \\
    \midrule
    Default            & activation-based & 0.3 & \textbf{74.82}      & 0.965 & 16.6 / 30.8 / 17.6 / 35.1 \\
    Random partition   & random partition & 0.3 & 73.95 ($-$0.87)     & 0.504 & 0.2 / 0.0 / 52.9 / 46.9 \\
    w/o resp.\ loss    & activation-based & 0.0 & 73.34 ($-$1.48)     & 0.491 & 0.0 / 0.0 / 57.4 / 42.6 \\
    \bottomrule
  \end{tabular}
\end{table*}

The two ablations expose complementary roles of the activation-based
channel partition and the routing-responsibility loss. Randomizing the
channel partition while keeping every other component of the
initialization pipeline intact makes the two L1 specializers functionally similar: their
output channels are now drawn uniformly from the same pool, so neither specializer develops a cluster-conditional representation that the L1 router can exploit. The L1 router consequently collapses to a single
branch and the
network degenerates into a two-leaf tree, with a Top-1 drop of
\(0.87\,\)pp on ImageNet-1K. Setting \(\lambda_{\mathrm{resp}} = 0\)
removes the explicit incentive for the L1 router to use both
branches: \(A_1\) and \(A_2\) each receive 0\,\% of validation
samples, and the accuracy drops by \(1.48\,\)pp.
\section{Discussion}\label{sec:discussion}

\subsection{Position on the Active-Parameter / Accuracy Trade-off}\label{sec:discussion:position}

$\Sigma$B-Net occupies a distinct operating point on the multi-axis
efficiency Pareto front, dominating the active-parameter axis while offering only moderate FLOPs reduction (Sections~\ref{sec:results:main}--\ref{sec:results:dynamic}).
The active-parameter advantage is structural rather than incidental: static
structured pruning removes filters permanently to compress both axes,
and sample-level dynamic computation gates entire blocks to skip
compute, whereas $\Sigma$B-Net reduces the active-parameter footprint
via per-inference path selection over a hierarchical binary tree,
with the dense total parameter count preserved.
DecisioNet~\cite{gottlieb2023decisionet} is the structurally closest
prior work, but its supervised label-tree construction is restricted
to convolutional backbones.

\subsection{Implications for Memory-Bound Deployment}\label{sec:discussion:edge}
The distinction between FLOPs and active parameters is central in the memory-bound regime. FLOPs measure arithmetic work, whereas \(P_{\mathrm{active}}\) measures the amount of model state that must be resident on-chip or transferred from off-chip memory for a single inference. When the dense model does not fit in on-chip memory and inference is performed at batch size one, weights cannot be amortized across a large batch, and off-chip weight traffic can dominate latency even when arithmetic units are underutilized. Thus, reducing
\(P_{\mathrm{active}}\) targets a different bottleneck from conventional FLOPs reduction.

The active-parameter advantage translates analytically into reduced per-inference
DRAM read traffic on memory-constrained accelerators, where on-chip memory cannot hold the dense network, and per-inference latency is
governed by off-chip weight transfer under the batch-size-one regime
typical of edge inference~\cite{pope2023inference}. Because only the
backbone, routers, and a single leaf execute, the transferred volume
scales with $P_{\mathrm{active}}$; an active-parameter reduction of approximately
60\,\% therefore reduces analytical off-chip transfer by approximately
60\,\%, corresponding under FP32 storage to roughly 41\,MB per
inference for the ResNet-50 $\Sigma$B-Net versus 102\,MB for the
dense baseline. Unlike static structured pruning, which achieves a
comparable reduction only at the cost of permanently lowered
capacity~\cite{hooker2019selective}, $\Sigma$B-Net decouples
per-inference memory traffic from the total parameter count.

Realizing this analytical reduction as wall-clock latency improvement is non-trivial: naive on-demand fetching after the router's argmax decision breaks the prefetch--compute overlap that hides memory latency, and selective weight transfer requires runtime support together with a branch-aligned memory layout that preserves burst-transfer efficiency. These implementation aspects are deferred to future work (Section~\ref{sec:discussion:future}). The present work isolates the algorithmic contribution of
active-parameter reduction from hardware-specific runtime
co-design, which we intentionally leave outside the scope of
this paper.

\subsection{Limitations}\label{sec:discussion:limitations}

Three limitations point to directions of future work. First, a non-trivial accuracy gap remains on ImageNet-1K (Section~\ref{sec:results:main}); a plausible cause is that the four-leaf structure does not fully absorb the intra-class variability of 1000 classes. Second, the modality scope does not
yet cover Vision Transformer~\cite{dosovitskiy2021vit}
architectures, which a framework-level generality claim would
ultimately require. Third, the reductions reported here are analytical; empirical wall-clock validation on memory-constrained accelerators is left for future work.

\subsection{Future Work}\label{sec:discussion:future}

Three directions follow from the limitations above. First, extension
to Vision Transformer backbones via a hybrid approach---dense
self-attention with hierarchically restructured feed-forward
sub-layers---would broaden modality coverage; recent analytical
FFN-to-MoE conversion~\cite{pei2026analytical} offers a flat,
training-free counterpart against which a hierarchical fine-tuned
variant could be compared. Second, the fixed $(2, 4)$ tree
configuration could be replaced with a data-driven choice via
activation-clustering criteria (e.g.\ silhouette or gap statistics)
or small-scale architecture search, better fitting larger label spaces and asymmetric input distributions. Third, hardware-level
realization on memory-constrained accelerators is required to convert the analytical traffic reduction into wall-clock latency
improvement, involving benchmarking on accelerators with explicit
on-chip / off-chip hierarchies, runtime support for selective weight
prefetching aligned with router decisions, and quantization-aware leaf placement to maximize on-chip residency.

\section{Conclusion}\label{sec:conclusion}
We introduced $\Sigma$B-Method, a framework that converts a
pretrained dense network into a hierarchical binary tree executed at
inference as a single hard top-1 path. Across CIFAR-100 / ResNet-50,
ImageNet-1K / ResNet-50, and ModelNet40 / PointNet++, $\Sigma$B-Net
reduces the per-inference active-parameter footprint by 58--60\,\%
while preserving classification accuracy within 1.72\,pp of the dense
baseline. By retaining the full dense parameter set, $\Sigma$B-Method decouples per-inference memory traffic from the total parameter count, occupying a distinct operating point relative to static
pruning and sample-level dynamic computation, and the cross-modal
evaluation substantiates a framework-level claim.

\bibliographystyle{IEEEtran}
\bibliography{IEEEabrv,references}

\begin{IEEEbiography}[{\includegraphics[width=1in,height=1.25in,clip,keepaspectratio]{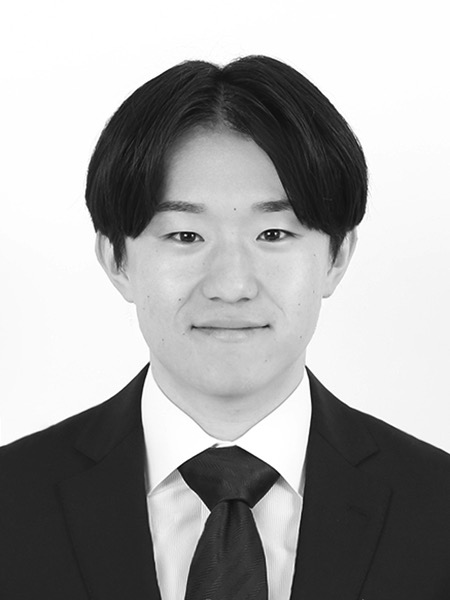}}]{KOHGA TANAKA}
received the B.E. degree from Keio University, Japan, in 2024, where
he is currently pursuing the master's degree. His research interests
include lightweight dynamic-inference frameworks, active-parameter
reduction, and edge AI.
\end{IEEEbiography}

\begin{IEEEbiography}[{\includegraphics[width=1in,height=1.25in,clip,keepaspectratio]{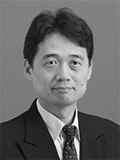}}]{HIROAKI NISHI}
He has been a Researcher with the Real World
Computing Partnership, since 1999, and with the Central Research
Laboratory, Hitachi Ltd., since 2002. He has been a Professor with
Keio University, since 2014. He is the Chair of IEEE P21451-1-6 and
IEEE P2992; and a member of IEEE 1451 Families, IEEE P2668, and
IEEE P2805. He was also a member of the ITU-T Focus Group on Smart
Sustainable Cities WG2. He is a member of several committees
established by the Ministry of Internal Affairs and Communications.
The main theme of his current research is a total network system
that includes the development of hardware and software architecture.
\end{IEEEbiography}

\end{document}